\documentclass[letterpaper]{article} 
\usepackage{aaai2026}  
\usepackage{times}  
\usepackage{helvet}  
\usepackage{courier}  
\usepackage[hyphens]{url}  
\usepackage{graphicx} 
\usepackage{natbib}  
\usepackage{caption} 
\usepackage{algorithm}
\usepackage{algorithmic}

\usepackage{newfloat}
\usepackage{listings}
\DeclareCaptionStyle{ruled}{labelfont=normalfont,labelsep=colon,strut=off} 
\floatstyle{ruled}
\newfloat{listing}{tb}{lst}{}
\floatname{listing}{Listing}
\usepackage{amssymb}

\usepackage{array}
\usepackage{multirow}
\usepackage{diagbox}
\usepackage{tablefootnote}
\usepackage{soul}
\usepackage{pifont}
\usepackage{amsmath,bm}
\usepackage{amsfonts}       
\usepackage{mathtools}
\usepackage{xcolor}

\usepackage{relsize}
\usepackage{booktabs} 
\usepackage{arydshln}
\usepackage{subfig}

\newcolumntype{C}{>{\centering\arraybackslash}p{1.3cm}}

\usepackage{times}
\usepackage{helvet}
\usepackage{courier}
\usepackage{xcolor}

\newcounter{checksubsection}
\newcounter{checkitem}[checksubsection]

\title{Echoless Label-Based Pre-computation for Memory-Efficient \\ Heterogeneous Graph Learning}

\author{
    Jun Hu\textsuperscript{\rm 1}, Shangheng Chen\textsuperscript{\rm 2}, Yufei He\textsuperscript{\rm 1}, Yuan Li\textsuperscript{\rm 1}, Bryan Hooi\textsuperscript{\rm 1}, Bingsheng He\textsuperscript{\rm 1}
}
\affiliations{
    \textsuperscript{\rm 1}National University of Singapore\\
    \textsuperscript{\rm 2}Institute of Automation, CAS\\
    jun.hu@nus.edu.sg, chenshangheng23@mails.ucas.ac.cn, \{yufei.he, li.yuan\}@u.nus.edu, \{dcsbhk, dcsheb\}@nus.edu.sg

}

\usepackage{bibentry}

\begin{document}

\maketitle

\begin{abstract}

Heterogeneous Graph Neural Networks (HGNNs) are widely used for deep learning on heterogeneous graphs. 
Typical end-to-end HGNNs require repetitive message passing during training, limiting efficiency for large-scale real-world graphs. 
Pre-computation-based HGNNs address this by performing message passing only once during preprocessing, collecting neighbor information into regular-shaped tensors, which enables efficient mini-batch training.
Label-based pre-computation methods collect neighbors' label information but suffer from training label leakage, where a node's own label information propagates back to itself during multi-hop message passing---the echo effect.
Existing mitigation strategies are memory-inefficient on large graphs or suffer from compatibility issues with advanced message passing methods.
We propose \textbf{Echoless Label-based Pre-computation (Echoless-LP)}, which eliminates training label leakage with Partition-Focused Echoless Propagation (PFEP). 
PFEP partitions target nodes and performs echoless propagation, where nodes in each partition collect label information only from neighbors in other partitions, avoiding echo while remaining memory-efficient and compatible with any message passing method. 
We also introduce an Asymmetric Partitioning Scheme (APS) and a PostAdjust mechanism to address information loss from partitioning and distributional shifts across partitions.
Experiments on public datasets demonstrate that Echoless-LP achieves superior performance and maintains memory efficiency compared to baselines.

\end{abstract}


\begin{links}
    \link{Code}{https://github.com/CrawlScript/Echoless-LP}
\end{links}

\section{Introduction}

\textbf{Heterogeneous graphs} contain multiple types of vertices and edges, unlike their homogeneous counterparts, making them well-suited for modeling complex real-world applications.
\textbf{Heterogeneous Graph Neural Networks (HGNNs)} emerge as powerful tools for deep learning on heterogeneous graphs, with applications spanning various domains such as social recommendation systems~\cite{DBLP:conf/mm/ZhouS23} and research classification~\cite{9950622,DBLP:conf/kdd/HuDWCS20}.

Typical HGNNs perform \textbf{end-to-end training} on graphs, with representative methods such as RGCN~\cite{schlichtkrull2018modeling} and HAN~\cite{DBLP:conf/www/WangJSWYCY19}.
During each forward pass, neighbors within K-hops are aggregated through message passing~\cite{kipf2016semi,velivckovic2018graph,hamilton2017inductive}. 
On heterogeneous graphs, neighbor aggregation is complicated and may involve multiple relation types, diverse node types, and complex meta-path traversals. 
This \textbf{repetitive message passing} occupies over 90\% of total training time~\cite{10643347}, causing scalability bottlenecks on large graphs.

\begin{figure}[!tp]
\centering
\includegraphics[width=2.8in]{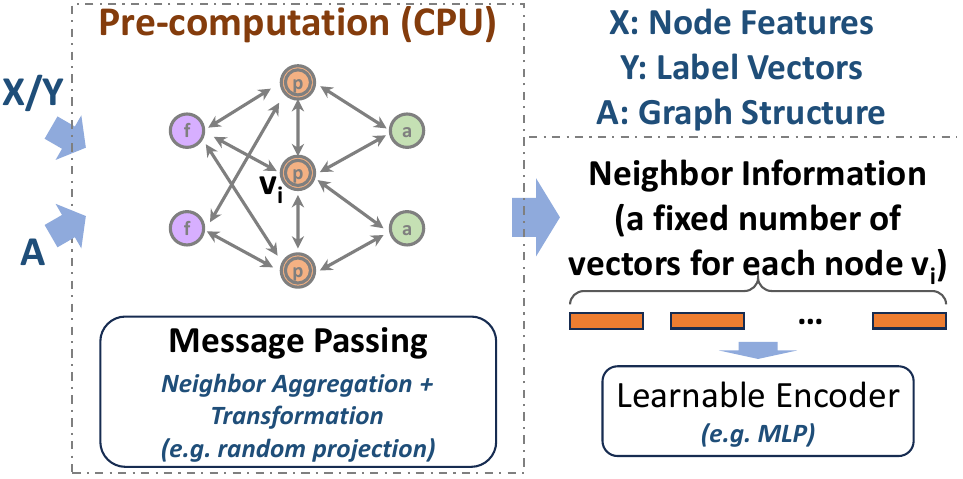}
\vspace{-1mm}
\caption{
Feature/label-based pre-computation.
}
\label{fig:intro_example_label_pre}
\vspace{-5mm}
\end{figure}

\begin{figure}[!t]
\centering
\subfloat[
Input/Output of target nodes.
]{
\includegraphics[width=0.49\linewidth]{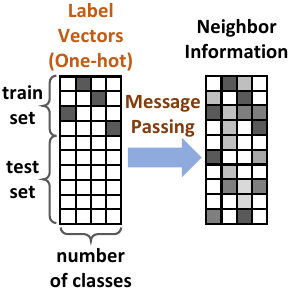}
\label{fig:intro_example_input_output} 
}
\subfloat[
Examples of echo.
]{
\includegraphics[width=0.51\linewidth]{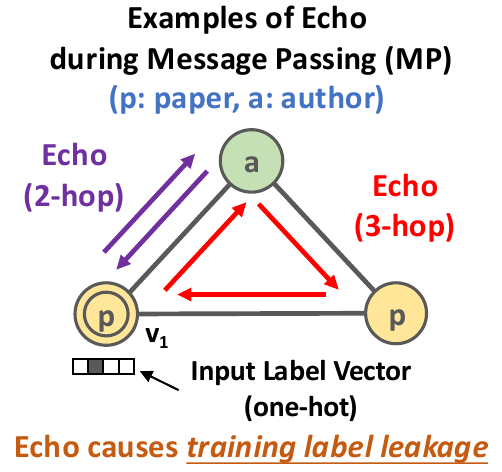}
\label{fig:intro_example_echo} 
}
\vspace{-1mm}
\caption{
$v_1$ is also a multi-hop neighbor of itself, and the 2-hop MP and 3-hop MP propagate $v_1$’s own label back to itself (\textbf{echo}), causing \textbf{training label leakage} for $v_1$.
}
\label{fig:intro_example} 
\vspace{-5mm}
\end{figure}

\textbf{Pre-computation for Efficient HGNNs.}
To address the efficiency limitations caused by repetitive message passing, pre-computation-based HGNNs emerge as a practical solution that only requires one-time message passing in preprocessing, and requires no graph operations during training. 
As shown in Figure~\ref{fig:intro_example_label_pre}, these methods operate in two stages: 
(1) a pre-computation stage that performs one-time message passing on CPUs to collect neighbor information within $K$ hops, generating a fixed number of vectors for each node to form regular-shaped tensors, and 
(2) a training stage where the encoder (e.g., MLP) takes the pre-computed tensors as input and performs training without any graph operations, thus enabling efficient mini-batch training. 
Depending on the input used for message passing during pre-computation, we distinguish between Feature-based Pre-computation (using node features $X$) and \textbf{Label-based Pre-computation} (using label vectors $Y$).
Figure~\ref{fig:intro_example_input_output} shows an example of $Y \in \{0,1\}^{N \times C}$, where training nodes use one-hot vectors and other nodes use zero vectors, and $N$ and $C$ are the number of nodes and classes, respectively.

Label-based pre-computation methods suffer from \textbf{training label leakage}, where a node's own label information is propagated back to itself during multi-hop message passing, as illustrated in Figure~\ref{fig:intro_example_echo}.
We define this phenomenon as the \textbf{echo effect}.
Echo causes a node's own label information to leak into the collected neighbor information (i.e., the node's representations).
During training, the learnable encoder may become reliant on this leaked self-label information, which is only available for training nodes but not for test nodes, leading to poor generalization performance.

\begin{figure}[!tp]
\centering
\includegraphics[width=2.7in]{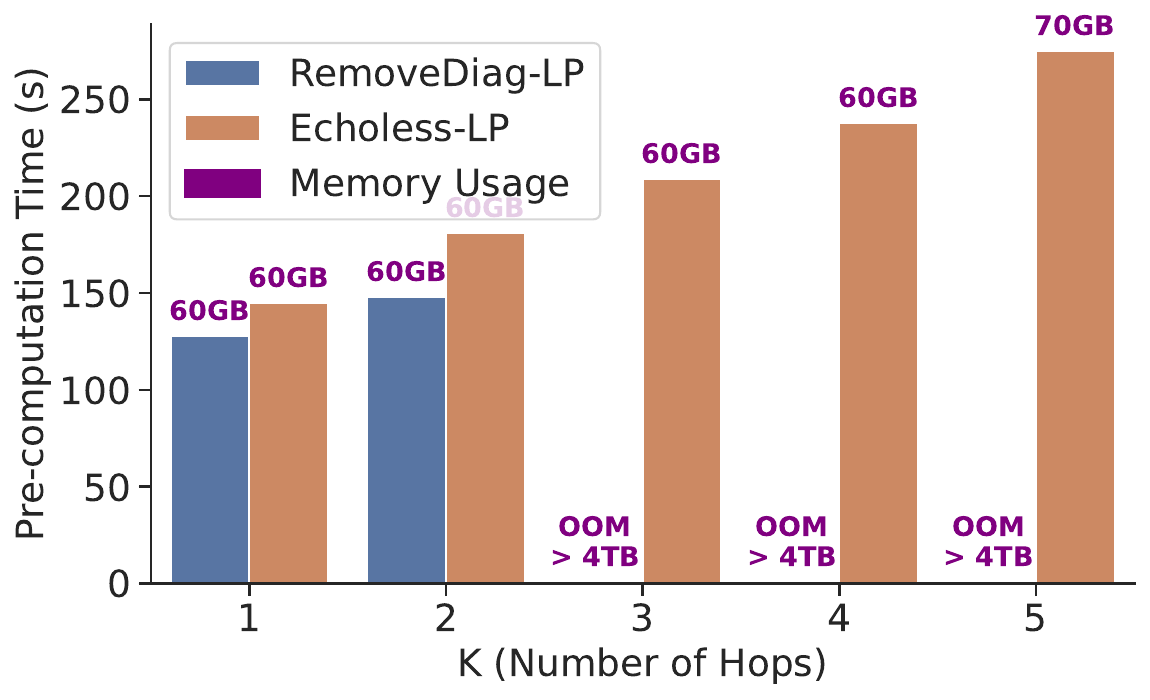}
\vspace{-2mm}
\caption{Memory usage (purple text) vs.\ number of hops $K$ for label‑based pre‑computation on OAG‑Venue (million‑scale).
Although Echoless‑LP incurs a modest increase in pre‑computation time (y‑axis), it remains memory‑efficient for $K>2$, whereas RemoveDiag‑LP (SOTA) runs out of memory (OOM).}
\label{fig:intro_hop_time_memory}
\vspace{-4mm}
\end{figure}

\textbf{Limitations of Existing Mitigation Strategies.}
Several methods have been proposed to address the echo effect in label-based pre-computation, such as high-hop filtering and propagation adjustment.
(1) Last Residual Connection~\cite{zhang2022graph} (referred to as LastResidual-LP in this paper) attempts to reduce label leakage by emphasizing neighbor information from higher-hop neighborhoods, where echo is generally less severe. 
However, echo can still occur at higher hops, so this approach \textbf{only partially alleviates} training label leakage and cannot eliminate it entirely.
(2) RemoveDiag-LP~\cite{yang2023simple} (a name we adopt for clarity) can avoid echo for linear message passing of the form $A_K A_{K-1} \dots A_1Y$, with each $A_i$ being a normalized adjacency matrix at hop $i$.
RemoveDiag-LP computes the entire multi-hop propagation as $\tilde{A} = A_K A_{K-1} \dots A_1 \in \mathbb{R}^{N \times N}$ and then removes the diagonal for message passing: $(\tilde{A} - \mathrm{diag}(\tilde{A}))Y$, where the diagonal removal effectively blocks echo. 
This changes the efficient computation order $A_K (A_{K-1} \dots (A_1Y)\dots)$, and the diagonal computation may involve the explicit construction of $\tilde{A}$, which may become increasingly dense when $K > 2$, leading to terabyte-scale memory usage for million-node graphs and making the method \textbf{memory-inefficient} (see Figure~\ref{fig:intro_hop_time_memory}).
Furthermore, as mentioned before, RemoveDiag-LP is designed upon linear message passing of the form $A_K A_{K-1} \dots A_1Y$, resulting in \textbf{compatibility issues} with more advanced message passing such as RpHGNN, which involve complex operations like feature normalization.

We propose \textbf{Echoless Label-based Pre-computation (Echoless-LP)}, a memory-efficient and compatible framework that eliminates training label leakage in label-based pre-computation on large graphs.
Echoless partitions target nodes into several partitions, and introduces Partition-Focused Echoless Propagation (PFEP).
PFEP performs echoless propagation via partition masks to collect neighbor information for one partition at a time, where each node can only collect information from neighbors
in other partitions, avoiding echo.
PFEP is \textbf{compatible with any message passing method} used in pre-computation, and message passing proceeds as usual without modification, avoiding costly memory overhead and ensuring \textbf{memory efficiency}.  
We also introduce an Asymmetric Partitioning Scheme (APS) and a \textbf{PostAdjust mechanism} to address potential information loss from partitioning and distributional shifts across partitions, respectively.
In experiments, we show that Echoless-LP integrates seamlessly with various message passing methods for pre-computation, achieves state-of-the-art (SOTA) performance, and efficiently supports high-hop propagation on large graphs (see Figure~\ref{fig:intro_hop_time_memory}).

Our main contributions are as follows:
\begin{itemize}
    \item We propose Echoless Label-based Pre-computation (Echoless-LP), which eliminates training label leakage caused by \textbf{echo} with a Partition-Focused Echoless Propagation (PFEP) component.
    PFEP is \textbf{compatible with any message passing method}, and message passing proceeds as usual without modification, avoiding costly memory overhead and ensuring \textbf{memory efficiency}.  
    \item We introduce an Asymmetric Partitioning Scheme (APS) and a PostAdjust mechanism to address information loss from partitioning and distributional shifts across partitions, respectively.
    \item Experiments on public datasets demonstrate that Echoless-LP achieves superior performance and maintains memory efficiency.
\end{itemize}

\section{Related Work}

\subsection{Heterogeneous Graph Neural Networks}

HGNNs leverage diverse node and edge types (relations) to capture rich node semantics.
RGCN~\cite{schlichtkrull2018modeling} extends GCN~\cite{kipf2016semi} and handles each relation type separately during message passing.
HAN~\cite{DBLP:conf/www/WangJSWYCY19} and MAGNN~\cite{DBLP:conf/www/0004ZMK20} consider message passing over meta-paths~\cite{DBLP:series/synthesis/2012Sun}, with hierarchical attention and intra/inter-metapath aggregation, respectively.
GTN~\cite{yun2019graph} enables automatic meta-path learning.
RSHN uses coarsened line graphs to capture the semantics of relations~\cite{zhu2019relation}.
HetSANN~\cite{hong2020attention} and HGT~\cite{DBLP:conf/www/HuDWS20} use attention-based approaches with type-aware attention layers and type-specific Transformers, respectively.
Simple-HGN~\cite{lv2021we} extends GAT~\cite{velivckovic2018graph} to consider relations for neighbor attention.
HINormer~\cite{DBLP:conf/www/MaoLLS23} proposes to use Transformers with local structure encoders and heterogeneous relation encoders.

\subsection{Pre-computation-based Heterogeneous Graph Neural Networks}

Pre-computation employs one-time CPU message passing to collect neighbor information, avoiding costly graph operations during training.
SIGN~\cite{sign_icml_grl2020} collects neighbor information at each hop separately for homogeneous graphs and then encodes it via MLPs.
NARS~\cite{yu2020scalable} samples relation-based subgraphs for heterogeneous graphs and applies SIGN on them.
SeHGNN~\cite{yang2023simple} collects neighbor information via meta-paths. 
RpHGNN~\cite{10643347} collects fine-grained multi-hop neighbor information, with complex operations (e.g., random projection, feature normalization) in message passing to ensure efficiency.
Given a specific pre-computation method, we can perform feature-based or label-based pre-computation, depending on whether we use node features or label information for message passing.

Label-based pre-computation may suffer from training label leakage caused by echo, and existing research mainly focuses on mitigation strategies.
Last Residual Connection~\cite{zhang2022graph} (LastResidual-LP) attempts to reduce label leakage by emphasizing higher-hop neighbors where echo is less severe, but only partially alleviates the problem as echo still occurs at higher hops.
RemoveDiag-LP~\cite{yang2023simple} removes diagonal elements from multi-hop adjacency matrices to block echo for linear message passing, but this approach is memory-inefficient for large graphs and incompatible with complex operations like feature normalization used in advanced methods.

Different from existing methods, our Echoless-LP avoids training label leakage caused by echo via echoless partition-focused propagation, and is compatible with any message passing method and memory-efficient on large graphs.

\section{Preliminary}

\textbf{Target Node Type (Target Type).} It is common to focus on only one node type in heterogeneous graph learning tasks.
For example, in an academic graph with papers and authors, the task may focus on classifying papers.
We refer to this specified type as the target node type (target type).
The nodes of the target type are defined as \textbf{target nodes}.

\textbf{Linear Message Passing.} Linear message passing refers to message passing of the form $A_K A_{K-1} \dots A_1 Y$, where each $A_i$ is a normalized adjacency matrix at hop $i$.
$\tilde{A} = A_K A_{K-1} \dots A_1$ is the entire \textbf{multi-hop propagation matrix}, which is not required to be explicitly computed for message passing due to efficiency issues.

\textbf{Efficiency of Diagonal Computation of Multi-hop Propagation Matrix}.
The SOTA method RemoveDiag-LP relies on diagonal removal.
(1) When the number of hops $K$ is 2, $\tilde{A} = A_2 A_1$, there is a trick to efficiently compute the diagonal without explicitly computing  $\tilde{A}$: $\text{diag}(\tilde{A}) = (A_2 \odot A_1^T) \mathbf{1}$, where $\odot$ denotes element-wise multiplication and $\mathbf{1}$ is a vector of ones. 
Using sparse matrix operations, the time and space complexity is $\mathcal{O}(E+N)$, where $N$ and $E$ are the number of nodes and edges, respectively.
(2) However, when $K > 2$, this trick does not apply, and we may have to explicitly compute the potentially dense matrix $\tilde{A} \in \mathbb{R}^{N \times N}$, which can reach terabyte-scale memory usage for million-node graphs, resulting in prohibitive memory overhead.

\begin{figure*}[!t]
\centering
\includegraphics[width=0.85\linewidth]{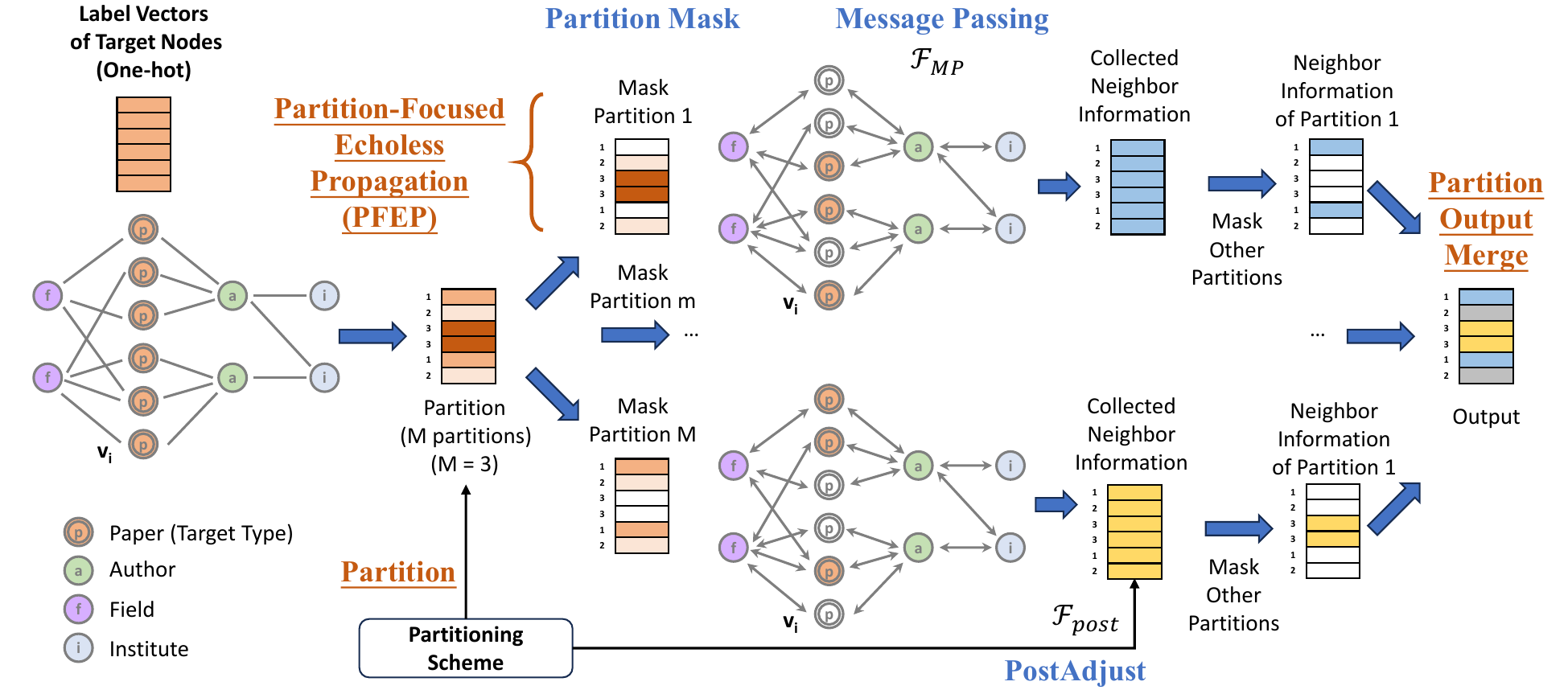} 
\vspace{-2mm}
\caption{Overall Framework of Echoless Label-Based Pre-computation (Echoless-LP).}
\label{fig:example_echo} 
\vspace{-2mm}
\end{figure*}

\begin{figure}[!t]
\vspace{-3mm}
\centering
\subfloat[
Uniform Random Partitioning.
]{
\includegraphics[width=0.5\linewidth]{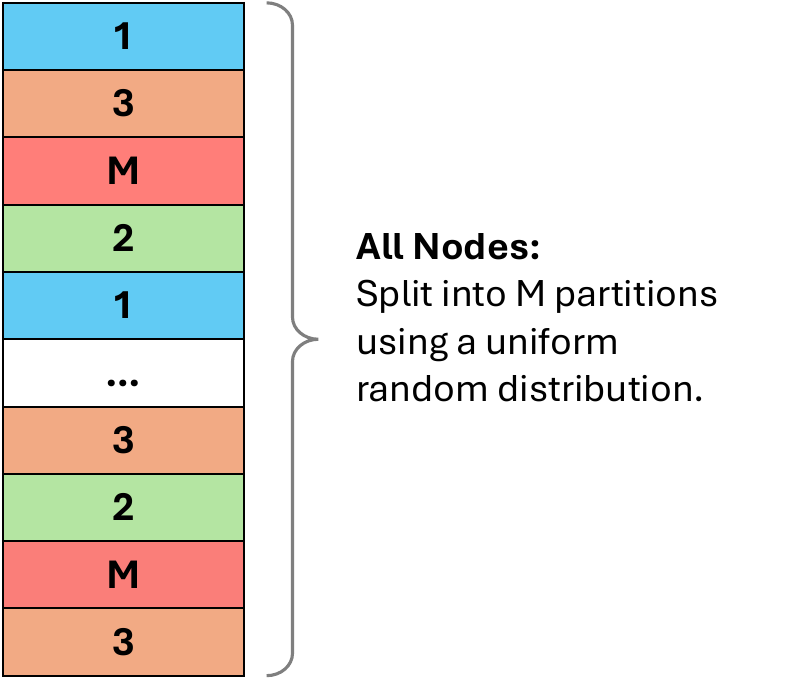}
\label{fig:partition_scheme_uniform}
}
\subfloat[
Asymmetric Partitioning.
]{
\includegraphics[width=0.5\linewidth]{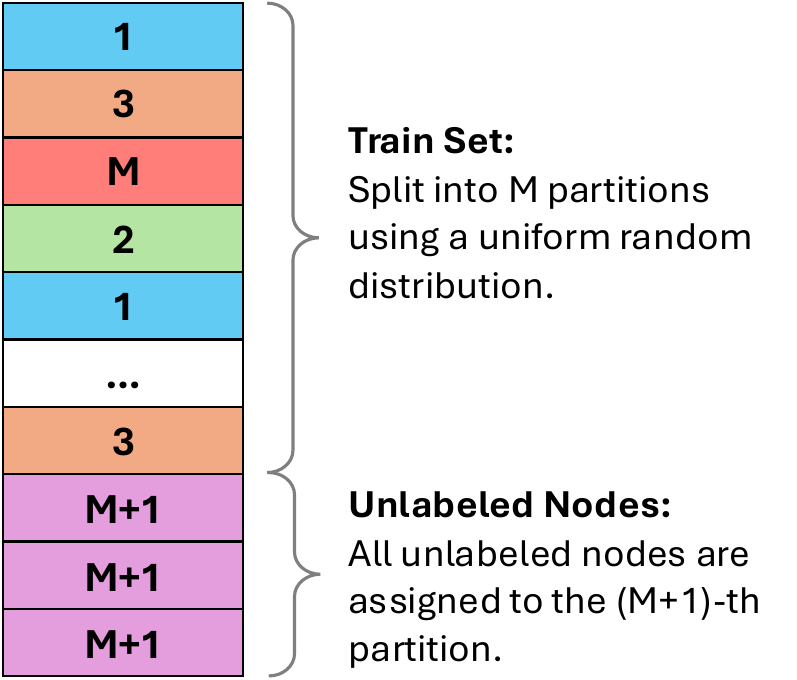}
\label{fig:partition_scheme_aps}
}
\vspace{-1mm}
\caption{
Partitioning Schemes.
}
\label{fig:partition_schemes} 
\vspace{-4mm}
\end{figure}

\section{Method}

In this section, we introduce our Echoless-LP framework.

\subsection{Echoless Label-based Pre-computation}

We propose Echoless Label-based Pre-computation (Echoless-LP), which avoids training label leakage caused by echo. 
The core idea of Echoless-LP is masking a node's input label vector to zero when collecting the node's neighbor information, which completely avoids the node's label information being propagated to itself (echo). 
For efficiency, we implement Echoless-LP with three key operations: 
\begin{itemize}
    \item \textbf{Partition}: Divide target nodes into multiple partitions using a certain partitioning scheme.

    \item \textbf{Partition-Focused Echoless Propagation (PFEP)}: Perform echoless propagation via partition masks to collect neighbor information for one partition at a time, where nodes in each partition collect label information only from neighbors in other partitions, avoiding echo.
    \item \textbf{Partition Output Merge}: Combine the results from all partitions to produce the final node representations.
\end{itemize}

Formally, given a heterogeneous graph $G$, let $V_t$ denote the set of target nodes to classify with $|V_t| = N$. 
We have target node features $X \in \mathbb{R}^{N \times d}$ and label vectors $Y \in \mathbb{R}^{N \times C}$, where $d$ is the feature dimension and $C$ is the number of classes. 
The three components operate as below:

\textbf{Partition.} We divide the target nodes $V_t$ into $M$ disjoint partitions $\{P_1, P_2, \ldots, P_M\}$ where:
\begin{equation}
\small
\bigcup_{i=1}^M P_i = V_t, \quad P_i \cap P_j = \emptyset \text{ for } i \neq j
\end{equation}
The partitioning scheme aims to minimize information loss by considering the graph structure and label distribution. Each partition $P_i$ contains a subset of target nodes that will be processed together in the subsequent propagation step. We discuss our partitioning scheme in the next subsection.

\textbf{Partition-Focused Echoless Propagation (PFEP).} For each partition $P_i$, we define a partition mask $\mathcal{M}_i \in \{0,1\}^{|V_t|}$ where:
\begin{equation}
\small
\mathcal{M}_i[j] = \begin{cases} 
1 & \text{if target node } j \in P_i \\
0 & \text{otherwise}
\end{cases}
\end{equation}
This mask ensures that only nodes in the current partition have their representations updated. We then apply a message propagation function that produces intermediate representations. 
Given a message passing operator $\mathcal{F}_{\text{MP}}$ used for feature-based pre-computation, which collects neighbor information as $H_i = \mathcal{F}_{\text{MP}}(X, G)$, we adapt it for PFEP by replacing the input node features with masked label vectors:
\begin{equation}
H_i = \mathcal{F}_{\text{MP}}(\text{diag}(\mathbf{1} - \mathcal{M}_i) Y, G)
\label{eq:mask_mp}
\end{equation}
Here, while message passing occurs across the entire graph $G$, the mask $(\mathbf{1} - \mathcal{M}_i)$ prevents label information of nodes in the current partition $P_i$ from being used when collecting neighbor information, achieving echoless propagation.

To address potential distribution shifts between partitions, we apply a \textbf{PostAdjust} mechanism $\mathcal{F}_{\text{post}}$ to each partition's output, which adjusts the output distribution across partitions as $\tilde{H}_i = \mathcal{F}_{\text{post}}(H_i)$.
The specific form of $\mathcal{F}_{\text{post}}$ depends on the partitioning scheme. 
In the next subsection, we introduce the $\mathcal{F}_{\text{post}}$ designed for our specific partitioning scheme.

\textbf{Partition Output Merge.} After PFEP, we obtain adjusted representations $\{\tilde{H}_1, \tilde{H}_2, \ldots, \tilde{H}_M\}$ from all $M$ partitions, which are then merged to produce the final target node representations as $H_{\text{label}} = \sum_{i=1}^M \text{diag}(\mathcal{M}_i) \tilde{H}_i$, where $\text{diag}(\mathcal{M}_i) \tilde{H}_i$ selects only the representations of nodes in partition $P_i$, and the summation combines the partition-specific representations into a unified output.

\subsection{Asymmetric Partitioning Scheme (APS)}

The performance of our Echoless-LP framework is affected by the design of the partitioning scheme.
A naive approach would uniformly and randomly partition all target nodes into $M$ partitions (see Figure~\ref{fig:partition_scheme_uniform}), but this may cause information loss.
Specifically, under uniform random partitioning, when processing partition $P_i$, we mask all nodes in that partition. For any node $v \in P_i$ collecting neighbor information, each of its neighbors has probability $1/M$ of also being in $P_i$ and thus having their labels masked, causing information loss.

To alleviate the information loss caused by partitioning, we propose an \textbf{Asymmetric Partitioning Scheme (APS)} that ensures unlabeled nodes retain all neighbor label information without information loss (see Figure~\ref{fig:partition_scheme_aps}).
Given labeled target nodes $V_{\text{train}} \subseteq V_t$ and unlabeled target nodes $V_{\text{unlabeled}} \subseteq V_t$, APS creates $M+1$ partitions:
\begin{itemize}
    \item \textbf{Training nodes:} Uniformly and randomly partitioned into $M$ disjoint partitions $\{P_1, \ldots, P_M\}$ where $\bigcup_{i=1}^{M} P_i = V_{\text{train}}$ and each $P_i$ contains approximately $|V_{\text{train}}|/M$ nodes.
    \item \textbf{Unlabeled nodes (validation/test):} Assigned to a dedicated partition $P_{M+1} = V_{\text{unlabeled}}$.
\end{itemize}
With APS, when collecting neighbor information for unlabeled nodes, only the nodes in $P_{M+1}$ (unlabeled nodes only) are masked, ensuring no neighbor label information is lost.

\textbf{PostAdjust in APS.} For APS, the PostAdjust mechanism is implemented as renormalization to alleviate distribution shifts. 
The asymmetry in APS creates distribution shifts where unlabeled nodes may accumulate neighbor information of different magnitude compared to training nodes.

To alleviate this shift, we first augment the input by creating $\overline{Y} = [\mathbf{1}_{\text{train}} \mid Y] \in \mathbb{R}^{N \times (C+1)}$ where $\mathbf{1}_{\text{train}} \in \{0,1\}^N$ indicates training nodes. 
The indicator column tracks retained neighbor information after masking.
After message passing $H_i = \mathcal{F}_{\text{MP}}(\text{diag}(\mathbf{1} - \mathcal{M}_i) \overline{Y}, G)$, we obtain $[\mathbf{r}_i \mid \overline{H}_{i}] = H_i$ where $\mathbf{r}_i \in \mathbb{R}^{N \times 1}$ contains the retention ratios. We normalize each row as follows:
%
\begin{equation}
\small
\tilde{H}_i = \operatorname{diag}\left( \frac{ \max(\mathbf{r}) }{ \mathbf{r}_i } \right) \overline{H}_i
\end{equation}
where $\mathbf{r} = \sum_{i=1}^{M} \text{diag} (\mathcal{M}_i) \mathbf{r}_i \in \mathbb{R}^{N \times 1}$ is the merged retention vector.
The division is element-wise, and the global maximum retention ratio $\max(\mathbf{r}) \in \mathbb{R}$ ensures that all node representations are scaled to the same reference level.

\subsection{Integration with Feature-based Pre-computation}

Our framework operates alongside existing feature-based pre-computation methods (e.g., RpHGNN)---which we term the \textbf{backbone}---using their message passing and learnable encoders for both features and labels.

\textbf{Message Passing.} Feature-based pre-computation methods usually apply $K^{\text{feat}}$ message passing operations $\{\mathcal{F}_{\text{MP}}^{(1)}, \mathcal{F}_{\text{MP}}^{(2)}, \ldots, \mathcal{F}_{\text{MP}}^{(K^{\text{feat}})}\}$ to collect $K^{\text{feat}}$ tensors:
\begin{equation}
\small
H_{\text{feat}}^{(k)} = \mathcal{F}_{\text{MP}}^{(k)}(X, G), \quad k = 1, 2, \ldots, K^{\text{feat}} 
\end{equation}
where $\mathcal{F}_{\text{MP}}^{(k)}$ differs in message passing hyperparameter $k$.
For simplicity, we use the hop index as the hyperparameter $k$, which is supported by most backbones.
For label-based pre-computation, we adopt the same message passing operations (but with a different number $K$) to collect $K$ tensors:
\begin{equation}
\small
H_{\text{label}}^{(k)} = \text{Echoless-LP}(Y, G, \mathcal{F}_{\text{MP}}^{(k)}), \quad k = 1, 2, \ldots, K
\end{equation}

\textbf{Learnable Encoder.} The collected tensors for features and labels are combined and processed by the backbone's encoder (e.g., hierarchical MLPs) for final classification:
\begin{equation}
\small
\hat{Y} = \text{Encoder}(\{H_{\text{feat}}^{(1)}, \ldots, H_{\text{feat}}^{(K^{\text{feat}})}, H_{\text{label}}^{(1)}, \ldots, H_{\text{label}}^{(K)}\})
\end{equation}

\subsection{Complexity Analysis}

Echoless-LP's complexity depends on the integrated backbone's message passing space/time complexity $\mathcal{O}(\mathcal{F}_{\text{MP}})$.

\textbf{Space Complexity.} 
Since message passing on each partition can be performed independently, the space complexity is $\mathcal{O}(\mathcal{F}_{\text{MP}}) + \mathcal{O}(NC)$, matching the backbone’s complexity, with an additional $\mathcal{O}(NC)$ term that accounts for storing intermediate label vectors during partition masking and merging. 
Thus, our method remains memory-efficient.

\textbf{Time Complexity.} The overall time complexity of Echoless-LP is $\mathcal{O}(M \cdot \mathcal{O}(\mathcal{F}_{\text{MP}})) + \mathcal{O}(NC)$.
The dominant $\mathcal{O}(M \cdot \mathcal{O}(\mathcal{F}_{\text{MP}}))$ term comes from performing message passing on $M$ partitions separately by PFEP, while the $\mathcal{O}(NC)$ term accounts for PostAdjust, partition masking and merging on label vectors. Our experiments show that usually a small $M$ of 2-4 is sufficient, keeping the overhead modest.

\section{Experiments}

In this section, we conduct node classification experiments on public datasets to verify the effectiveness of our methods.
All experiments are conducted on a Linux machine with dual Intel(R) Xeon(R) E5-2690 v4 CPUs, 128 GB RAM, and a GeForce GTX 1080 Ti (11 GB).

\subsection{Datasets and Task}
We evaluate Echoless-LP on various public datasets widely adopted for heterogeneous graph node classification research.
The datasets include three small-scale datasets: DBLP, IMDB, and Freebase from the HGB benchmark~\cite{lv2021we}, and three large-scale datasets: OGBN-MAG~\cite{hu2020open}, OAG-Venue, and OAG-L1-Field~\cite{sinha2015overview,tang2008arnetminer,zhang2019oag}, which have been used in large-scale HGNN research~\cite{yu2020scalable,10643347}.
Detailed statistics for all datasets are provided in Table~\ref{tab:datasets_statistics}.

Each dataset specifies a target node type, and the task is to classify nodes of this type (target nodes).
We use the fixed train/valid/test splits officially provided by the datasets.

\subsection{Baselines}
We compare our proposed Echoless-LP framework with two main categories of baselines: (1) HGNN-based approaches, including homogeneous GNNs (GCN~\cite{kipf2016semi}, GAT~\cite{velivckovic2018graph}, GraphSAGE~\cite{hamilton2017inductive}), heterogeneous GNNs (RGCN~\cite{schlichtkrull2018modeling}, HAN~\cite{DBLP:conf/www/WangJSWYCY19}, GTN~\cite{yun2019graph}, RSHN~\cite{zhu2019relation}, HetGNN~\cite{zhang2019heterogeneous}, MAGNN~\cite{DBLP:conf/www/0004ZMK20}, HetSANN~\cite{hong2020attention}, HGT~\cite{DBLP:conf/www/HuDWS20}, Simple-HGN~\cite{lv2021we}, HINormer~\cite{mao2023hinormer}), and pre-computation-based HGNNs (NARS~\cite{yu2020scalable}, SeHGNN~\cite{yang2023simple}, RpHGNN~\cite{10643347}) that reduce computational overhead by performing graph propagation only once before training; and (2) label-based pre-computation methods that leverage label information as plug-in modules for pre-computation HGNNs.

The label-based pre-computation baselines include LastResidual-LP~\cite{zhang2022graph} and RemoveDiag-LP~\cite{yang2023simple}.
All of them are integrated with backbones NARS, SeHGNN, and RpHGNN.
One exception is that RemoveDiag-LP only supports linear message passing and is therefore incompatible with the RpHGNN backbone, which has complex message passing operations.
We use italic font to denote the backbone, e.g., the method name Echoless-LP (\textit{RpHGNN}) denotes Echoless-LP with RpHGNN as the backbone.

\begin{table}[!tp]
\centering
\setlength{\tabcolsep}{1.4mm} 
\scalebox{0.83}{
\small
\begin{tabular}{>{\centering\arraybackslash}m{1.9cm}|c| >{\centering\arraybackslash}m{7mm}|c|>{\centering\arraybackslash}m{7mm}|>{\centering\arraybackslash}m{1cm}|>{\centering\arraybackslash}m{1cm}}\hline

Dataset      & Vertices   & Vertex Types & Edges      & Edge Types & Target Type       & Target Classes   \\\hline
DBLP         & 26,128     &      4       & 239,566    & 3          & $\mathtt{author}$ & 4   \\\hline
IMDB         & 21,420     &      4       & 86,642     & 3          & $\mathtt{movie}$  & 5   \\\hline
Freebase     & 180,098    &      8       & 1,057,688  & 36         & $\mathtt{book}$   & 7   \\\hline
OGBN-MAG     & 1,939,743  &      4       & 21,111,007 & 4          & $\mathtt{paper}$ & 349 \\\hline
OAG-Venue    & 1,116,162  &      5       & 13,985,692 & 15         & $\mathtt{paper}$ & 3505 \\\hline
OAG-L1-Field & 1,116,163  &      5       & 13,016,549 & 15         & $\mathtt{paper}$ & 275  \\\hline
\end{tabular}
}
\vspace{-1mm}
\caption{Statistics of datasets.}
\label{tab:datasets_statistics}
\vspace{-4mm}
\end{table}

\begin{table*}[!tp]
  \centering
\setlength{\tabcolsep}{0.8mm} 
\scalebox{0.78}{
\small

\begin{tabular}{l | c c c c c c | c c c c c}\hline

 & \multicolumn{6}{c|}{Small} & \multicolumn{5}{c}{Large} \\ \hline
               & \multicolumn{2}{ c }{DBLP}      & \multicolumn{2}{c}{IMDB}               & \multicolumn{2}{c|}{Freebase} & OGBN-MAG & \multicolumn{2}{c}{OAG-Venue} & \multicolumn{2}{c}{OAG-L1-Field}\\\hline

                     & Macro-F1       & Micro-F1       & Macro-F1       & Micro-F1       & Macro-F1       & Micro-F1       & Accuracy   & NDCG      & MRR  & NDCG      & MRR \\\hline

GCN                       & 90.84$\pm$0.32 & 91.47$\pm$0.34 & 57.88$\pm$1.18 & 64.82$\pm$0.64 & 27.84$\pm$3.13 & 60.23$\pm$0.92 & OOM & OOM & OOM & OOM & OOM   \\
GAT                       & 93.83$\pm$0.27 & 93.39$\pm$0.30 & 58.94$\pm$1.35 & 64.86$\pm$0.43 & 40.74$\pm$2.58 & 65.26$\pm$0.80 & OOM & OOM & OOM & OOM & OOM   \\

GraphSAGE                  & 92.53$\pm$0.53 & 93.06$\pm$0.47 & 58.36$\pm$0.70 & 61.92$\pm$0.72 & 44.17$\pm$1.14 & 61.71$\pm$0.53 & 46.46$\pm$0.60 & 40.32$\pm$0.91 & 23.05$\pm$0.80 & 52.12$\pm$0.09 & 54.09$\pm$0.08 \\\hline

RGCN                      & 91.52$\pm$0.50 & 92.07$\pm$0.50 & 58.85$\pm$0.26 & 62.05$\pm$0.15 & 46.78$\pm$0.77 & 58.33$\pm$1.57 & 48.11$\pm$0.48 & 48.93$\pm$0.26 & 31.51$\pm$0.20 & 85.91$\pm$0.10 & 84.92$\pm$0.23\\
HAN                        & 91.67$\pm$0.49 & 92.05$\pm$0.62 & 57.74$\pm$0.96 & 64.63$\pm$0.58 & 21.31$\pm$1.68 & 54.77$\pm$1.40 & OOM & OOM & OOM & OOM & OOM   \\
GTN                       & 93.52$\pm$0.55 & 93.97$\pm$0.54 & 60.47$\pm$0.98 & 65.14$\pm$0.45 & OOM            & OOM            & OOM & OOM & OOM & OOM & OOM   \\
RSHN                      & 93.34$\pm$0.58 & 93.81$\pm$0.55 & 59.85$\pm$3.21 & 64.22$\pm$1.03 & OOM            & OOM            & OOM & OOM & OOM & OOM & OOM   \\
HetGNN                   & 91.76$\pm$0.43 & 92.33$\pm$0.41 & 48.25$\pm$0.67 & 51.16$\pm$0.65 & OOM            & OOM            & OOM & OOM & OOM & OOM & OOM   \\
MAGNN                    & 93.28$\pm$0.51 & 93.76$\pm$0.45 & 56.49$\pm$3.20 & 64.67$\pm$1.67 & OOM            & OOM            & OOM & OOM & OOM & OOM & OOM  \\
HetSANN                  & 78.55$\pm$2.42 & 80.56$\pm$1.50 & 49.47$\pm$1.21 & 57.68$\pm$0.44 & OOM            & OOM            & OOM & OOM & OOM & OOM & OOM  \\
HGT                      & 93.01$\pm$0.23 & 93.49$\pm$0.25 & 63.00$\pm$1.19 & 67.20$\pm$0.57 & 29.28$\pm$2.52 & 60.51$\pm$1.16 & 46.78$\pm$0.42 & 47.31$\pm$0.32 & 29.82$\pm$0.33 & 84.13$\pm$0.37 & 82.16$\pm$0.89 \\

Simple-HGN               & 94.01$\pm$0.24 & 94.46$\pm$0.22 & 63.53$\pm$1.36 & 67.36$\pm$0.57 & 47.72$\pm$1.48 & 66.29$\pm$0.45 & OOM & OOM & OOM & OOM & OOM  \\

HINormer                  & 94.57$\pm$0.23 & 94.94$\pm$0.21 & 64.65$\pm$0.53 & 67.83$\pm$0.34 & 49.94$\pm$2.12 & 66.47$\pm$0.47 & OOM & OOM & OOM & OOM & OOM \\\hline

NARS                     & 94.18$\pm$0.47 & 94.61$\pm$0.42  & 63.51$\pm$0.68 & 66.18$\pm$0.70 & 49.98$\pm$1.77 & 63.26$\pm$1.26 &50.66$\pm$0.22 & 52.28$\pm$0.17 & 34.38$\pm$0.21 & 86.06$\pm$0.10 & 85.15$\pm$0.14\\

SeHGNN &  94.86$\pm$0.14 & 95.24$\pm$0.13 & 66.63$\pm$0.34 & 68.21$\pm$0.32 & 50.71$\pm$0.44 & 63.41$\pm$0.47 & 51.45$\pm$0.29 & 46.75$\pm$0.27 & 29.11$\pm$0.25 & 86.01$\pm$0.21 & 84.95$\pm$0.20 \\
RpHGNN & 95.23$\pm$0.31 & 95.55$\pm$0.29 & \underline{67.53$\pm$0.79} & 69.77$\pm$0.66  & 54.02$\pm$0.88 & 66.55$\pm$0.67 & 52.07$\pm$0.17 & 53.31$\pm$0.40 & 35.46$\pm$0.47 & \underline{87.80$\pm$0.06} & \underline{86.79$\pm$0.18} \\\hline

LastResidual-LP (\textit{NARS})                   & 94.30$\pm$0.51     & 94.70$\pm$0.43     & 63.98$\pm$0.33     & 66.30$\pm$0.49         & 50.18$\pm$2.48     & 63.92$\pm$0.84     & 52.28$\pm$0.25     & 53.03$\pm$0.20     & 35.44$\pm$0.17     & 86.21$\pm$0.11     & 85.30$\pm$0.14     \\
LastResidual-LP (\textit{SeHGNN}) 
                               & 92.51$\pm$0.47 & 93.08$\pm$0.37 &  63.18$\pm$0.98 & 65.43$\pm$0.67 &   44.74$\pm$2.35 & 57.82$\pm$3.01      &  47.59$\pm$0.54      &   42.02$\pm$0.13 & 24.67$\pm$0.13     &  83.98$\pm$0.06 & 82.30$\pm$0.10\\
LastResidual-LP (\textit{RpHGNN}) 
                                & 95.27$\pm$0.32 & 95.58$\pm$0.29 &  66.67$\pm$0.83 & 70.22$\pm$0.54 &  \underline{54.17$\pm$0.92} & 66.63$\pm$0.49      &   45.88$\pm$0.25    &   49.65$\pm$0.12 & 32.17$\pm$0.08     &  87.48$\pm$0.11     & 86.31$\pm$0.26 \\\hline

RemoveDiag-LP (\textit{NARS})         & 94.48$\pm$0.71 & 94.87$\pm$0.61 & 64.04$\pm$0.57 & 66.31$\pm$0.76 &    50.35$\pm$1.40     &  63.78$\pm$0.90         & 53.99$\pm$0.09 & 53.10$\pm$0.35     &  35.59$\pm$0.29     &  86.85$\pm$0.02 &  86.11$\pm$0.10 \\

RemoveDiag-LP (\textit{SeHGNN}) 
                              & 95.06$\pm$0.17  & 95.42$\pm$0.17 & 66.48$\pm$0.57 & 68.36$\pm$0.42  & 51.87$\pm$0.86  & 65.08$\pm$0.66    & 54.00$\pm$0.22 & 44.21$\pm$0.28 & 27.02$\pm$0.22 & 86.57$\pm$0.04 & 85.13$\pm$0.23 \\

RemoveDiag-LP (\textit{RpHGNN}) 
                              & --  & -- & -- & -- & -- & -- & -- & -- & -- & -- & --  \\\hline

Echoless-LP (\textit{NARS})       & 95.18$\pm$0.29 & 95.49$\pm$0.28
     & 64.50$\pm$0.31 & 67.11$\pm$0.46  & 50.58$\pm$1.61 & 63.85$\pm$1.21 & \textbf{54.43$\pm$0.14}     & \underline{53.68$\pm$0.22} & \underline{36.16$\pm$0.27} & 86.89$\pm$0.07     & 86.16$\pm$0.11     \\

Echoless-LP (\textit{SeHGNN}) 
                              & \underline{95.31$\pm$0.23}  & \underline{95.59$\pm$0.22} & 67.08$\pm$0.53 & \underline{70.12$\pm$0.43} & 52.84$\pm$0.51 & \underline{66.95$\pm$0.30} & \underline{54.17$\pm$0.18} & 47.92$\pm$0.33 & 30.47$\pm$0.31 & 86.81$\pm$0.08 & 85.43$\pm$0.16 \\

Echoless-LP (\textit{RpHGNN}) 
                                & \textbf{95.39$\pm$0.26} & \textbf{95.70$\pm$0.25} & \textbf{67.99$\pm$0.54} & \textbf{70.84$\pm$0.39}    &  
                         \textbf{54.28$\pm$0.83} & \textbf{67.08$\pm$0.58} &   54.04$\pm$0.17   & \textbf{54.72$\pm$0.83} & \textbf{37.33$\pm$0.80} &  \textbf{88.11$\pm$0.04} & \textbf{87.28$\pm$0.06}\\\hline

\end{tabular}}
\vspace{-2mm}
\caption{Performance on Small and Large Datasets. Best results are in bold and second-best results are underlined. 
RemoveDiag-LP (\textit{RpHGNN}) results are marked as ``--'', since RemoveDiag-LP is incompatible with RpHGNN’s complex message passing.
}
\label{tab:performance}
  \vspace{-5mm}
  \end{table*}

\subsection{Evaluation Metrics and Parameter Settings}

Following existing work~\cite{yu2020scalable,yang2023simple,10643347}, we use different metrics for different datasets: Macro-F1 and Micro-F1 for DBLP, IMDB, and Freebase; accuracy for OGBN-MAG; and NDCG and MRR for OAG-Venue and OAG-L1-Field.

For all baselines except label-based pre-computation methods, we adopt existing results from the literature~\cite{lv2021we,10643347}.
For label-based pre-computation methods, we consider the integration of each method with three backbones, which is not covered by existing research.
We carefully implement these integrations using the official code of each label-based pre-computation method and backbone method.
To ensure robust evaluation, we perform each experiment using 10 different random seeds and report the average performance with standard deviation. %
Following prior work~\cite{yu2020scalable,yang2023simple}, we use early stopping based on validation performance to save the best model for evaluation.

For parameter settings, we consider them for both backbones and label-based pre-computation methods.
For the integrated backbones, we try our best to follow their official parameter settings, such as learning rate, optimizer, hidden size, and number of layers of encoders.
Note that the hyperparameters for each backbone remain unchanged when integrated with different label-based pre-computation methods, ensuring fair comparison among them.
For label-based pre-computation hyperparameters, the key hyperparameters tuned include the number of hops for label information ($K$) and the number of partitions ($M$), with $K \in [1, 8]$ and $M \in [2, 5]$.
Besides, for the learnable encoders (e.g., hierarchical MLP), we only consider the input dropout rate $d_{in}$ and hidden dropout rate related to the collected label information (not the feature information), both searched within $[0.0, 0.9]$.
All parameters are selected via performance on the validation set.

\subsection{Performance Analysis}

\begin{figure}[!t]
\centering
\includegraphics[width=3.3in]{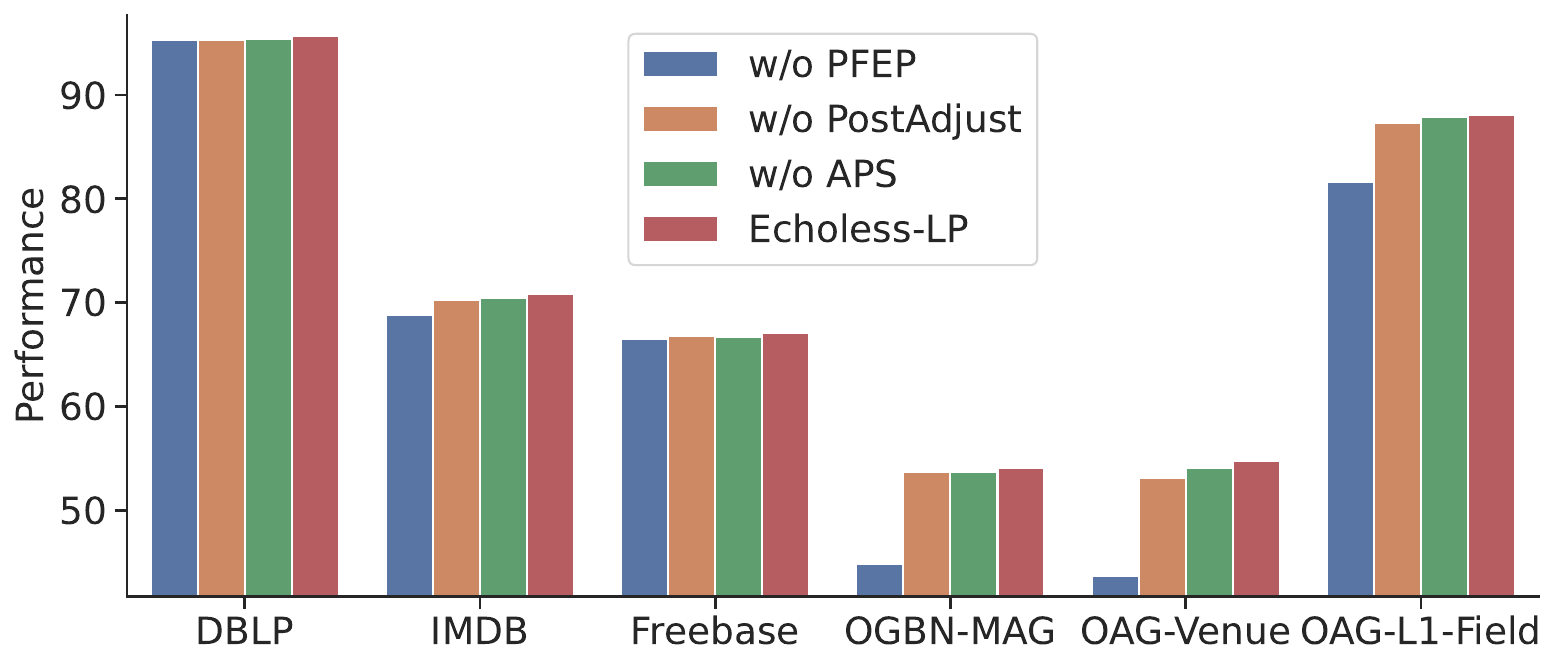}
\vspace{-6mm}
\caption{
Ablation study. We use F1-Micro for DBLP, IMDB, and Freebase; Accuracy for OGBN-MAG; and NDCG for OAG-Venue and OAG-L1-Field.
}
\label{fig:ablation_study}
\vspace{-3mm}
\end{figure}

Based on Table~\ref{tab:performance}, we have the following observations:
\begin{itemize}
\item Pre-computation-based GNNs, including NARS, SeHGNN, and RpHGNN, show overall superior performance and efficiency over end-to-end GNNs, showing the advantage of pre-computation-based GNNs.
\item By introducing labels as extra input data, the label-based pre-computation method LastResidual-LP cannot consistently improve the backbone's performance, outperforming its NARS backbone but underperforming its SeHGNN and RpHGNN backbones.
One reason for the performance drop is that LastResidual-LP only alleviates the training label leakage issue but does not address it completely.
Additionally, the severity of training label leakage may vary when using different message passing methods (provided by the backbone).
\item The label-based pre-computation method RemoveDiag-LP can avoid training label leakage and improves most backbones (NARS and SeHGNN), showing that avoiding training label leakage is necessary for effective learning. 
However, it has compatibility issues and cannot integrate with the most powerful backbone RpHGNN, as it only supports linear message passing and is thus incompatible with RpHGNN, which involves complex message passing operations like feature normalization.
\item Echoless-LP achieves the best performance and most of the second-best performance across datasets.
Both Echoless-LP and RemoveDiag-LP can avoid training label leakage. 
However, RemoveDiag-LP is limited to 2 hops on large graphs due to costly memory overhead, while Echoless-LP can utilize label information within $K$ hops ($K > 2$) without such overhead.
Additionally, unlike RemoveDiag-LP, which is limited to linear message passing, Echoless-LP is compatible with the powerful backbone RpHGNN with complex message passing operations. 
As a result, Echoless-LP achieves superior overall performance, demonstrating its effectiveness.

\end{itemize}

\subsection{Detailed Analysis}

\begin{figure}[!t]
\centering
\subfloat[
DBLP
]{
\includegraphics[width=0.5\linewidth]{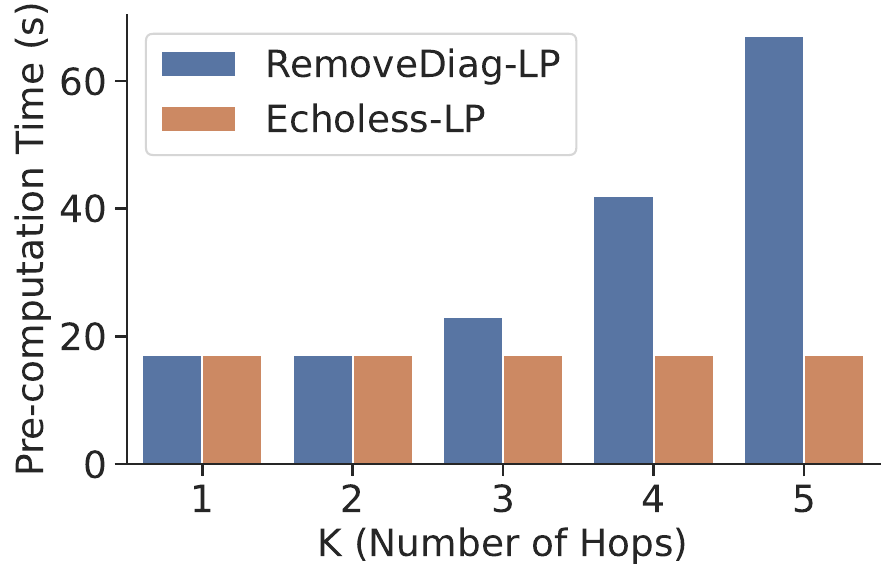}
}
\subfloat[
OGBN-MAG
]{
\includegraphics[width=0.5\linewidth]{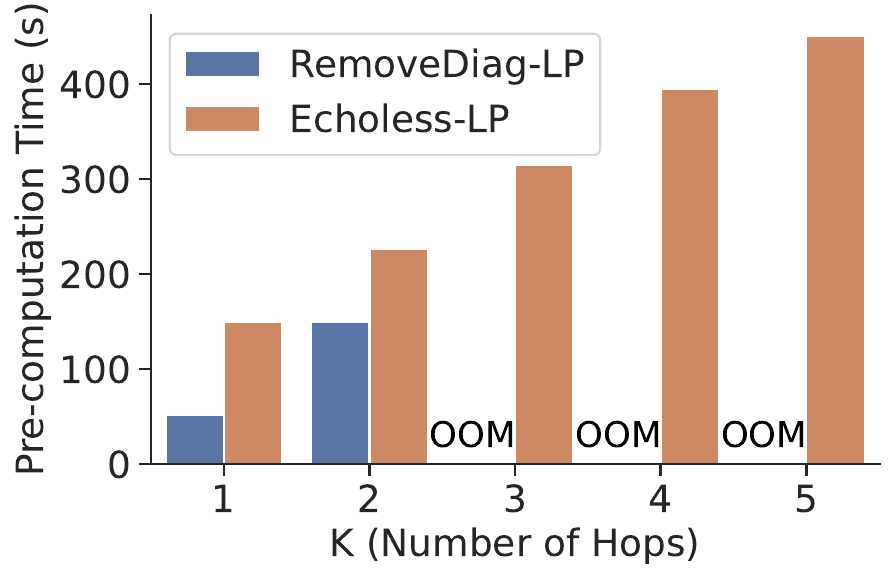}
}
\\
\subfloat[
OAG-Venue
]{
\includegraphics[width=0.5\linewidth]{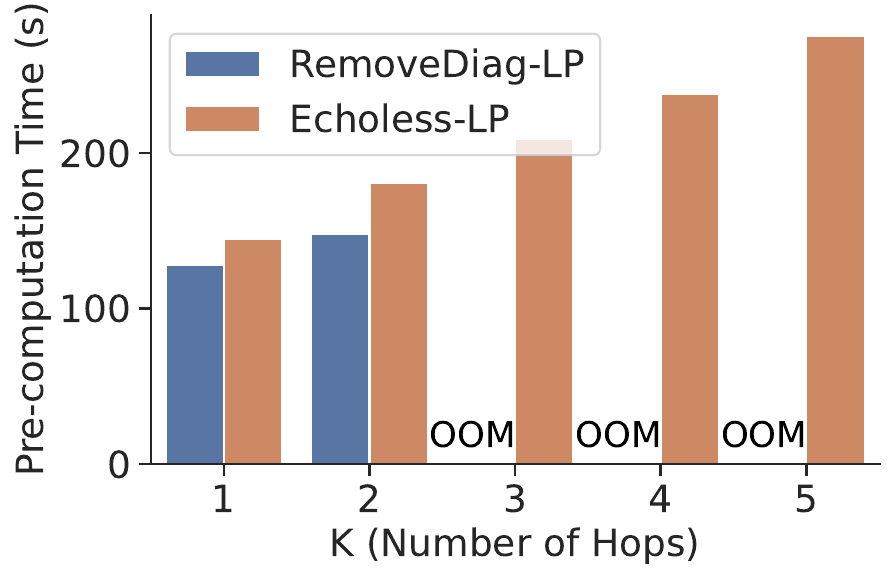}
}
\subfloat[
OAG-L1-Field
]{
\includegraphics[width=0.5\linewidth]{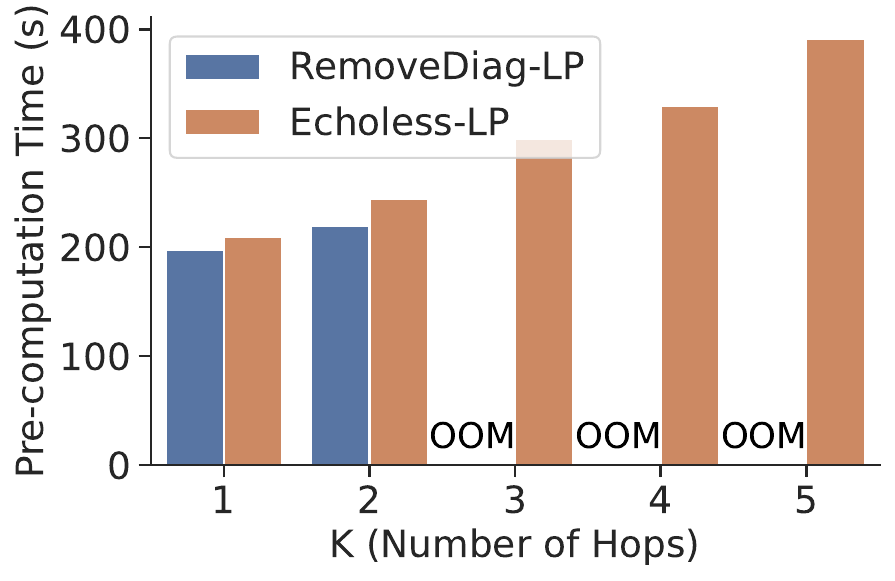}
}
\vspace{-2mm}
\caption{
Pre-computation time (s) vs. number of hops $K$.
}
\label{fig:result_pre_time_k} 
\vspace{-2mm}
\end{figure}

\begin{table}[!t]
\centering
\scalebox{0.8}{
\small
\begin{tabular}{l|c|c|c|c}
\hline
\multirow{2}{*}{$K$} & \multicolumn{2}{c|}{OGBN-MAG} & \multicolumn{2}{c}{OAG-Venue}  \\
\cline{2-5}
 & RemoveDiag-LP & Echoless-LP & RemoveDiag-LP & Echoless-LP \\
\hline
1 &  40GB      &   50GB &  60GB      &  60GB  \\
\hline
2 &  50GB      &   60GB &  60GB      &  60GB  \\
\hline
3 & OOM ($>$ 13TB) &   70GB &  OOM ($>$ 4TB) &  60GB \\
\hline
4 & OOM ($>$ 13TB) &   80GB & OOM ($>$ 4TB)  &  60GB  \\
\hline
5 & OOM ($>$ 13TB) &   90GB & OOM ($>$ 4TB)  &  70GB  \\
\hline
\end{tabular}
}
\vspace{-2mm}
\caption{Memory usage vs. number of hops $K$. Memory for non-OOM cases is rounded up to the nearest 10GB, and OOM cases show estimated memory requirements.}
\label{tab:result_pre_mem_k}
\vspace{-6mm}
\end{table}

\subsubsection{Ablation Study}
To verify the effectiveness of main components, we design several variants of our Echoless-LP: 
(1) \textit{Echoless-LP w/o PFEP} removes the Partition-Focused Echoless Propagation. Instead, it directly uses feature-based pre-computation methods to collect neighbor label information by replacing the input node features with label vectors.
(2) \textit{Echoless-LP w/o APS} uses uniform random partitioning scheme, instead of our Asymmetric Partitioning Scheme.
(3) \textit{Echoless-LP w/o PostAdjust} skips the PostAdjust operation after message passing.
Results in Figure~\ref{fig:ablation_study} compare their performance (using RpHGNN as the backbone).

Echoless-LP outperforms Echoless-LP w/o PFEP, benefiting from PFEP's ability to avoid training label leakage caused by echo.
Our method also achieves better results than Echoless-LP w/o APS, demonstrating that APS effectively reduces information loss caused by partitioning.
Additionally, Echoless-LP surpasses Echoless-LP w/o PostAdjust, confirming that alleviating distribution shifts between partitions improves performance.

\begin{figure}[!t]
\centering
\subfloat[
DBLP
]{
\includegraphics[width=0.5\linewidth]{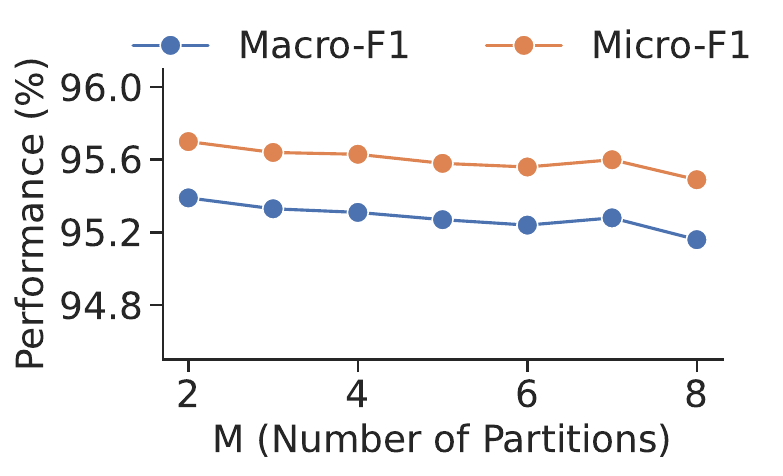}
}
\subfloat[
IMDB
]{
\includegraphics[width=0.5\linewidth]{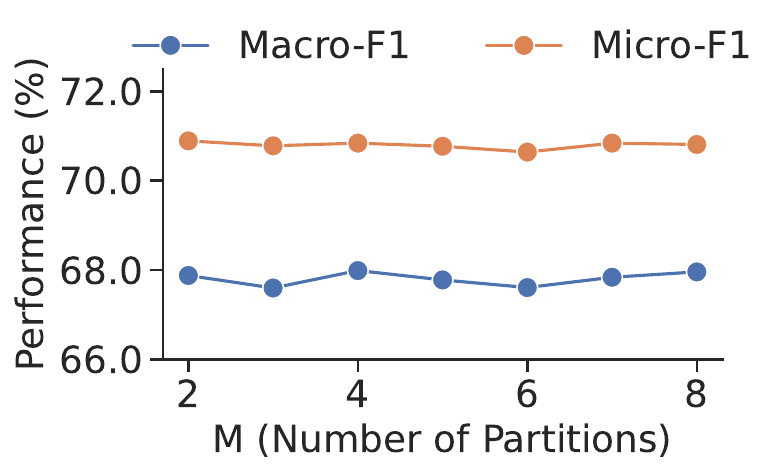}
}
\\
\subfloat[
OGBN-MAG
]{
\includegraphics[width=0.5\linewidth]{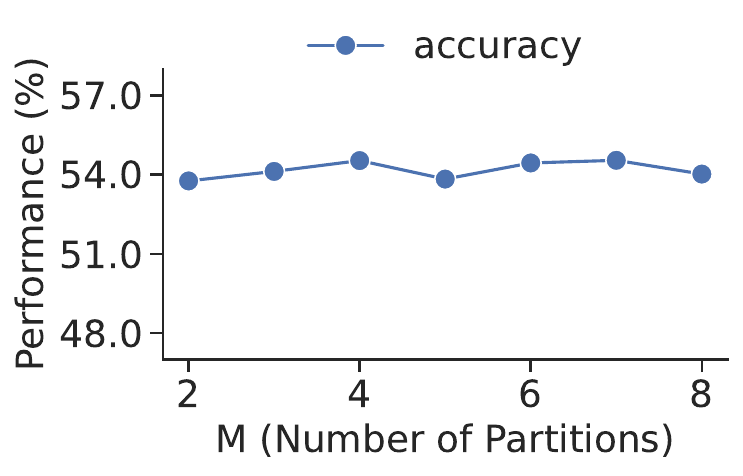}
}
\subfloat[
OAG-Venue
]{
\includegraphics[width=0.5\linewidth]{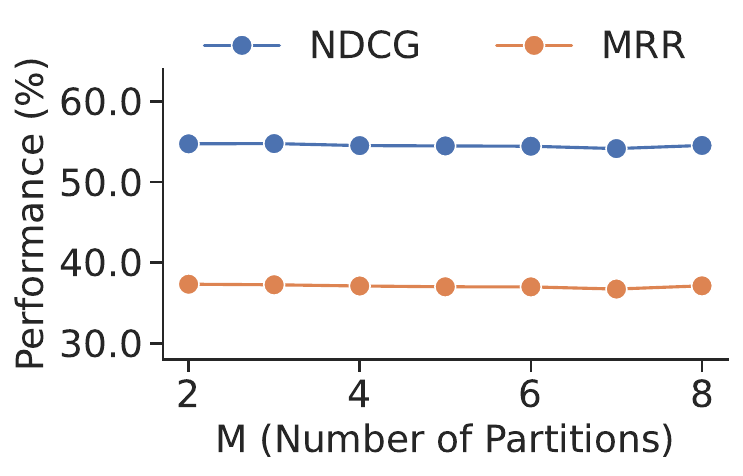}
}
\vspace{-2mm}
\caption{
Impact of the number of partitions $M$. 
}
\label{fig:impact_num_partitions} 
\vspace{-5mm}
\end{figure}

\subsubsection{Memory-Efficiency of Pre-computation}
We vary the number of hops $K$ and report pre-computation time in Figure~\ref{fig:result_pre_time_k} using NARS as the backbone.
While Echoless-LP costs moderately more time than RemoveDiag-LP when $K \leq 2$ due to separate message passing on $M$ partitions, it demonstrates clear efficiency advantages when $K > 3$.
RemoveDiag-LP becomes much slower on the small dataset (DBLP) and runs out of memory (OOM) on large datasets (OGBN-MAG, OAG-Venue, and OAG-L1-Field), whereas Echoless-LP maintains approximately linear time scaling.
This efficiency gap is explained by memory usage shown in Table~\ref{tab:result_pre_mem_k}: when $K > 2$, RemoveDiag-LP requires explicit computation of the multi-hop propagation matrix for diagonal removal, causing terabyte-level memory usage, while Echoless-LP avoids such costly overhead.

\subsubsection{Impact of Number of Partitions}

Although our APS eliminates information loss for unlabeled nodes (including test nodes), training nodes still experience information loss affected by the number of partitions $M$, where labels of any neighbor have a $1/M$ probability of being masked. 
Intuitively, a larger $M$ should reduce this information loss and improve performance.
However, results in Figure~\ref{fig:impact_num_partitions} show that a small $M$ (e.g., $M \leq 4$) is typically sufficient, with $M=2$ achieving optimal performance in most cases.
One potential explanation is that, for training nodes, the moderate information loss introduced by smaller $M$ values may act as a form of beneficial regularization, improving model robustness rather than hindering performance.

\section{Conclusion}
We propose Echoless Label-based Pre-computation (Echoless-LP) for memory-efficient heterogeneous graph learning.
It eliminates training label leakage caused by echo with Partition-Focused Echoless Propagation (PFEP).
PFEP is compatible with any message passing method, and message passing proceeds as usual without modification, avoiding costly memory overhead and ensuring memory efficiency.  
We introduce Asymmetric Partitioning Scheme (APS) and PostAdjust to address information loss from partitioning and distributional shifts across partitions, respectively.
Experiments show that Echoless-LP achieves superior performance and maintains memory efficiency.

\section{Acknowledgments}

This research is supported by the National Research Foundation, Singapore and Infocomm Media Development Authority under its Trust Tech Funding Initiative, and the Ministry of Education, Singapore, under the Academic Research Fund Tier 2 (FY2025) (Grant MOE-T2EP20124-0009). Any opinions, findings and conclusions or recommendations expressed in this material are those of the author(s) and do not reflect the views of National Research Foundation, Singapore and Infocomm Media Development Authority.

\bibliography{aaai2026}

\end{document}